\pdfoutput=1

\documentclass[11pt]{article}

\usepackage[final]{acl}

\usepackage{times}
\usepackage{latexsym}

\usepackage[T1]{fontenc}

\usepackage[utf8]{inputenc}

\usepackage{microtype}

\usepackage{inconsolata}

\usepackage{graphicx}
\usepackage{amsmath}

\usepackage{enumitem} 
\usepackage{booktabs}

\title{CasiMedicos-Arg: A Medical Question Answering Dataset \\Annotated with Explanatory Argumentative Structures}

\author{
Ekaterina Sviridova\textsuperscript{1}\thanks{Equal contribution} \quad
Anar Yeginbergen\textsuperscript{2}\footnotemark[1] \quad
Ainara Estarrona\textsuperscript{2} \\
\textbf{Elena Cabrio}\textsuperscript{1} \quad
\textbf{Serena Villata}\textsuperscript{1} \quad
\textbf{Rodrigo Agerri}\textsuperscript{2} \\
\textsuperscript{1}Université Côte d’Azur, CNRS, Inria, I3S, France\\
\textsuperscript{2}HiTZ Center - Ixa, University of the Basque Country UPV/EHU \\
\texttt{\{sviridova, cabrio, villata\}@i3s.unice.fr},\\
\texttt{\{anar.yeginbergen, ainara.estarrona, rodrigo.agerri\}@ehu.eus}\\
}

\begin{document}
\maketitle
\begin{abstract}
Explaining Artificial Intelligence (AI) decisions is a major challenge nowadays in AI, in particular when applied to sensitive scenarios like medicine and law. However, the need to explain the rationale behind decisions is a main issue also for human-based deliberation as it is important to justify \textit{why} a certain decision has been taken. Resident medical doctors for instance are required not only to provide a (possibly correct) diagnosis, but also to explain how they reached a certain conclusion. Developing new tools to aid residents to train their explanation skills is therefore a central objective of AI in education. In this paper, we follow this direction, and we present, to the best of our knowledge, the first multilingual dataset for Medical Question Answering where correct and incorrect diagnoses for a clinical case are enriched with a natural language explanation written by doctors. These explanations have been manually annotated with argument components (i.e., premise, claim) and argument relations (i.e., attack, support), resulting in the Multilingual CasiMedicos-Arg dataset which consists of 558 clinical cases in four languages (English, Spanish, French, Italian) with explanations, where we annotated 5021 claims, 2313 premises, 2431 support relations, and 1106 attack relations. We conclude by showing how competitive baselines perform over this challenging dataset for the argument mining task.
\end{abstract}

\section{Introduction} \label{sec1}

There is an increasingly large body of research on AI applied to the medical domain with the objective of developing technology to assist and support medical doctors in explaining their decisions or how they have reached a certain conclusion. For example, resident medical doctors preparing for licensing exams may get AI support to explain what and why is the treatment or diagnosis correct given some background information \cite{LLMMedEduc:23,goenaga2023explanatory}.  

A prominent example of this is the recent proliferation of Medical Question Answering (QA) datasets and benchmarks, in which the task often involves processing and acquiring relevant specialized medical knowledge to be able to answer a medical question based on the context provided by a clinical case \cite{singhal2023large,nori2023capabilities,xiong2024benchmarking}.

The development of Large Language Models (LLMs), both general purpose and specialized in the medical domain, has enabled rapid progress in Medical QA tasks which has led in turn to claims about LLMs being able to pass official medical exams such as the United States Medical Licensing Examination (USMLE) \cite{singhal2023towards,nori2023capabilities}. Thus, publicly available LLMs such as LLaMa \cite{touvron2023llama} or Mistral \cite{jiang2023mistral} and their respective medical-specific versions PMC-LLaMa \cite{wu2023pmcllama} and BioMistral \cite{labrak2024biomistral}, or proprietary models such as MedPaLM \cite{singhal2023towards} and GPT-4 \cite{nori2023capabilities}, to name but a few, have been reporting high-accuracy scores in a variety of Medical QA benchmarks\footnote{\url{https://huggingface.co/blog/leaderboard-medicalllm}}\cite{singhal2023large,singhal2023towards,xiong2024benchmarking}.

While these results constitute impressive progress, currently the Medical QA research field still presents a number of shortcomings. First, experimentation has been mostly focused on providing the correct answer in medical exams, usually in a multiple-choice setting. However, as doctors are also required to explain and argue about their predictions, research on Medical QA should also address the identification and generation of argumentative explanations. Unfortunately, and to the best of our knowledge, 
no Medical QA dataset currently includes correct and incorrect diagnoses enriched with annotated argumentative structures written by medical doctors. Second, the large majority of Medical QA benchmarks are available only in English \cite{singhal2023large,xiong2024benchmarking}, which makes it impossible to test the ability of current LLMs for Medical QA in other languages.

In this paper we address these issues by presenting CasiMedicos-Arg, the first Multilingual (English, French, Italian, Spanish) dataset for Medical QA with manually annotated gold explanatory argumentation about incorrect and correct predictions written by medical doctors. More specifically, we take the CasiMedicos corpus \cite{Agerri2023HiTZAntidoteAE,ALONSO2024102938} consisting of 558 documents with reference gold doctors' explanations, and enrich it with manual annotations for argument components (5021 claims and 2313 premises) and relations (2431 support and 1106 attack). This new resource will make it possible, for the first time, to research not only on Argument Mining but also on generative techniques to argue about and explain predictions in Medical QA settings. Finally, strong baselines on argument component detection, a challenging sequence labelling task, using encoder \cite{devlin-etal-2019-bert,he2021debertav3}, encoder-decoder \cite{medical-mt5} and decoder-only LLMs \cite{jiang2023mistral,touvron2023llama} demonstrate the validity of our annotated resource. Data, code and fine-tuned models are publicly available (\url{https://github.com/ixa-ehu/antidote-casimedicos}).

\section{Related Work} \label{sec2}

In this section, we will focus on reviewing datasets for Medical QA and on Explanatory Argumentation, the two features of our main contribution, CasiMedicos-Arg.

\subsection{Medical Question Answering}

Several of the most popular Medical QA datasets \cite{jin-etal-2019-pubmedqa,abacha2019overview,abacha2019bridging,jin2021disease,pal2022medmcqa} have been grouped into three multi-task English benchmarks, namely, MultiMedQA \cite{singhal2023large}, \textsc{MIRAGE} \cite{xiong2024benchmarking}, and the Open Medical-LLM Leaderboard \cite{Medical-LLM-Leaderboard}, with the aim of providing comprehensive experimental evaluation benchmarks of LLMs for Medical QA.

MultiMedQA includes MedQA \cite{jin2021disease}, MedMCQA \cite{pal2022medmcqa}, PubMedQA \cite{jin-etal-2019-pubmedqa}, LiveQA \cite{abacha2019overview}, MedicationQA \cite{abacha2019bridging}, MMLU clinical topics \cite{hendrycks2020measuring} and HealthSearchQA \cite{singhal2023large}. Except for the last one, all of them consist of a multiple-choice format and MedQA, MedMCQA and MMLU's source data come from licensing medical exams. In terms of size, MedQA includes almost 15K questions, MedMCQA 187K while the rest of them are of more moderate sizes, namely, 500 QA pairs in PubMedQA, around 1200 in MMLU, 738 in LiveQA and 674 in MedicationQA. 

While every dataset except MedQA and HealthSearchQA includes long form correct answers, they are not considered really usable for benchmarking LLMs because they were not optimally constructed as a \emph{ground-truth} by medical doctors or professional clinicians \cite{singhal2023large}.

The Open Medical-LLM Leaderboard also includes MedQA, MedMCQA, PubMedQA and MMLU clinical topics. General purpose LLMs such as GPT-4 \cite{nori2023capabilities}, PaLM \cite{Chowdhery2022PaLMSL}, LLaMa \cite{touvron2023llama} or Mistral \cite{jiang2023mistral} report high-accuracy scores on these Medical QA benchmarks, although recently a number of specialized LLMs for the medical domain sometimes appear with even stronger performances. Some popular models include Med-PaLM \cite{singhal2023large}, MedPaLM-2 \cite{singhal2023towards}, PMC-LLaMa \cite{wu2023pmcllama}, and more recently, BioMistral \cite{labrak2024biomistral}.

The \textsc{MIRAGE} benchmark includes subsets of MedQA, MedMCQA, PubMedQA, MMLU clinical topics and adds the BioASQ-YN dataset \cite{tsatsaronis2015overview} with the aim of evaluating Retrieval Augmented Generation (RAG) techniques for LLMs in Medical QA tasks. According to the authors, their \textsc{MedRAG} method not only helps to address the problem of hallucinated content by grounding the generation on specific contexts, but it also provides relevant up-to-date knowledge that may not be encoded in the LLM \cite{xiong2024benchmarking}. By employing \textsc{MedRAG}, they are able to clearly improve the zero-shot results of some of the tested LLMs, although the results for others are rather mixed.

Finally, MedExpQA \cite{ALONSO2024102938} presents the first multilingual Medical QA benchmark with reference gold explanations of incorrect and incorrect options written by Spanish medical doctors. The benchmark is translated into French, English and Spanish and it includes also RAG experiments. 
 
To summarize, no Medical QA dataset currently provides manually annotated gold argumentative explanations for both the correct and incorrect predictions. Furthermore, and with the exception of \citet{vilares2019head} and \citet{ALONSO2024102938}, they have been mostly developed for English, leaving a huge gap regarding the evaluation of LLMs in Medical QA for other languages. Motivated by this we present CasiMedicos-Arg, the first Medical QA dataset including gold reference explanations written by medical doctors \cite{ALONSO2024102938} which we manually annotated with argumentative structures, including argument components (premises and claims) and their relations (support and attack).

\subsection{Explanatory Argumentation in the Medical Domain} \label{subsec2.2}

Explanatory argumentation in natural language refers to the process of generating or analyzing explanations within argumentative texts. In recent years, natural language explanation generation has gained significant attention due to the advancements of generative models that are leveraged to develop specialized explanatory systems. The need for explanation generation is also driven by the predominant use of non-transparent algorithms which lack interpretability, thus being unsuitable for sensitive domains such as medical. 

\citet{camburu2018eSNLINL} tackle the task of explanation generation by introducing an extension of the Stanford Natural Language Inference (SNLI) dataset \citep{bowman2015large}, which includes a new layer of annotations providing explanations for the entailment, neutrality, or contradiction labels. The generation of these explanations is addressed with a bi-LSTM encoder trained on the new e-SNLI dataset. e-SNLI \citep{camburu2018eSNLINL} is also exploited to generate explanations for a NLI method, which first generates possible explanations for predicted labels (Label-specific Explanations) and then takes a final label decision \citep{kumar-talukdar-2020-nile}. The authors use GPT-2 \citep{radford2019language} for label-specific generation and classify explanations with RoBERTa \citep{liu2019roberta}. 

\citet{narang2020wt5} focus on generating complete explanations in natural language following a prediction step, utilizing a T5 model. The model is trained to predict both the label and the explanation. \citet{li2021you} also propose to generate explanations along with predicting NLI labels. The generation step is leveraged for the question-answering task exploiting domain-specific or commonsense knowledge, while the NLI step allows to predict relations between a premise and a hypothesis. \citet{kotonya-toni-2024-towards} propose a framework to rationalize explanations taking into account not only free-form explanations, but also argumentative explanations. Furthermore, authors provide metrics for explanation evaluation. 

In the medical domain, \citet{molinet2024} propose generating template-based explanations for medical QA tasks. Their system incorporates medical knowledge from the Human Phenotype Ontology, making the explanations more verifiable and sound for the medical domain. At the same time, quality assessment of medical explanations remains challenging, as the process of decision-making is not transparent. In this regard,  \citet{DBLP:conf/aiia/MarroCCV23} propose a new methodology to evaluate reasons of explanations in clinical texts.

Despite the extensive research proposing various approaches to generate explanations, these approaches are not grounded on any argumentation model. This is particularly important in sensitive domains like medicine, where sound and well-founded explanations are essential to justify the taken decision. Moreover, medical explanations require verified medical knowledge at their core, which the described methods lack, as discussed in~\citet{molinet2024}.

\section{CasiMedicos-Arg Annotation} \label{sec4}

The Spanish Ministry of Health yearly publishes the Resident Medical or \emph{M\'edico Interno Residente} (MIR) licensing exams including the correct answer. Every year the CasiMedicos MIR Project 2.0\footnote{\url{https://www.casimedicos.com/mir-2-0/}} takes the published exams by the ministry and provide gold explanatory arguments written by volunteer Spanish medical doctors to reason about the correct and incorrect options in the exam.

The Antidote CasiMedicos corpus consists of the original Spanish commented exams by the CasiMedicos doctors which were cleaned, structured and freely released for research purposes \cite{Agerri2023HiTZAntidoteAE}. The original Spanish data was automatically translated and manually revised into English, French, and Italian. The corpus includes 622 documents each with a short clinical case, the multiple-choice questions and the explanations written by medical doctors\footnote{\url{https://huggingface.co/datasets/HiTZ/casimedicos-exp}} and it has been used to setup the first multilingual benchmark for Medical QA \cite{ALONSO2024102938}.

In the rest of this section we describe the process of manually annotating argumentative structures in the raw Antidote CasiMedicos dataset.

\subsection{Argumentation Annotation Guidelines} \label{subsec4.1}


In line with the guidelines proposed by \citet{mayer2021enhancing} for Randomized Controlled Trials (RCT) annotation, we identify two main argument components: Claims and Premises, and their relations, Support and Attack. Furthermore, we also propose to annotate Markers and labels specific to the medical domain, namely, Disease, Treatment and Diagnostics. In the following, we define and describe each type of annotation.

A \textbf{Claim} is a concluding statement made by the author about the outcome of the study \citep{mayer2021enhancing}:

\begin{enumerate}
    \item \emph{The patient's presenting picture is presumably erythema nodosum. (CasiMedicos)}
    \item \emph{We propose immunotherapy with thymoglobulin and cyclosporine as a proper treatment. (CasiMedicos)}
\end{enumerate}

A \textbf{Premise} corresponds to an observation or measurement in the study, which supports or attacks another argument component, usually a Claim. It is important that they are observed facts, therefore, credible without further evidence \citep{mayer2021enhancing}:

\begin{enumerate}[resume]
    \item \emph{In addition, pancytopenia is not observed. (CasiMedicos)}
    \item \emph{What is important is that the eye that has received the blow does not go up, and therefore there is double vision in the superior gaze. (CasiMedicos)}
\end{enumerate}

Analyzing the CasiMedicos dataset, we found certain ambiguity between claims and premises. Thus, statements representing general medical knowledge about a disease, symptoms, or treatments must be annotated as claims. Although these statements may support or attack the main claim, they are not premises since they do not involve case-specific evidence but represent medical facts:

\begin{enumerate}[resume]
    \item \emph{[The patient's presenting picture is presumably erythema nodosum]. [About 10\% of cases of erythema nodosum are associated with inflammatory bowel disease, both ulcerative colitis and Crohn's disease]. [As mentioned, in most cases, erythema nodosum has a self-limited course]. [When associated with inflammatory bowel disease, erythema nodosum usually resolves with treatment of the intestinal flare, and recurs with disease recurrences. Local measures include elevation of the legs and bed rest]. (CasiMedicos)}
\end{enumerate}

Here the first statement in square brackets represents a claim that asserts the patient's diagnosis (\emph{erythema nodosum}). The following ones represent information about the diagnosis, its symptoms and its possible treatment. They are not based on the evidences given in the case, but on general medical knowledge available to the doctor. Therefore, these examples should be annotated as Claims.

Additionally, long statements with multiple self-contained pieces of evidence must be divided into single premises to differentiate their relations to specific claims. For example, a given evidence in a sentence may support a claim while others may attack it. To preserve these distinctions, such sentences should be split into independent premises.

As well as Claims and Premises we annotate \textbf{Markers} – discourse markers that are relevant for arguments as they help to identify the spans of argument components and the type of argumentative relations. In the following examples markers are written in bold:

\begin{enumerate}[resume]
    \item \emph{\textbf{Other causes} related to this picture are autoimmune diseases \textbf{leading to} transverse myelitis (Behcet's, FAS, SLE,...) \textbf{or} inflammatory diseases such as sarcoidosis, \textbf{although} our patient does not seem to meet the criteria for them. (CasiMedicos)}
    \item \emph{\textbf{Although} this usually gives a subacute \textbf{or} chronic picture. (CasiMedicos)}
\end{enumerate}

The possible answers proposed in the CasiMedicos multiple-choice options correspond to predicting a \textbf{Disease}, a \textbf{Treatment} or a \textbf{Diagnosis}. We decided to also annotate them as they help to identify the type of doctor's arguments (whether to look justification of a diagnosis or about a possible treatment) and the type of argumentative relations.

For advanced reasoning comprehension, we need to explore argumentative relations connecting argument components (claims and premises) and forming a structure of an argument \cite{mayer2021enhancing}. Here we provide the definitions of support and attack relations, as well as real examples illustrating them. 

\textbf{Support.} All statements or observations justifying the proposition of a target argument component are considered as supportive \cite{mayer2021enhancing}:

\begin{enumerate}[resume]
    \item \emph{In the examination there is a clear dissociation with thermoalgesic anesthesia and preservation of arthrokinetic and vibratory. [1] Reflexes are normal, neither abolished nor exalted. [2] In addition, the rest of the examination is strictly normal. [3] \textbf{With all this I believe that the correct answer is 5, that is a syringomyelic lesion, whose initial characteristic is the sensitive dissociation with anesthesia for the thermoalgesic and conservation of the posterior chordal}. (CasiMedicos)}
\end{enumerate}

This example provides premises (in italics) that justify a claim (bold) which they are related to. The supportive nature is highlighted by the marker \emph{With all this I believe...}.

\textbf{Attack.} An argument component is attacking another one if (i) it contradicts the proposition of a target component or (ii) it undercuts its implicit assumption of significance or relevance, for example, stating that the observations related to a target component are not significant or not relevant \cite{mayer2021enhancing}:

\begin{enumerate}[resume]
    \item \emph{\textbf{It might be tempting to answer 3 Fracture of the superior wall of the orbit with entrapment of the superior rectus muscle}. However, muscles trapped in a fracture do not automatically lose their muscular action. (CasiMedicos)}
    \item \emph{The palpebral hematoma and hyposphagma (subconjunctival hemorrhage) does not give us the key data. (CasiMedicos)}
\end{enumerate}

These examples represent premises (in italics) which either contradict their claims (bold) in Example 9 or which are not considered significant to justify or reject target components (Example 10). 

\subsection{CasiMedicos Real Case Example}

In this section we describe a real CasiMedicos case annotated with argument components – Premises (in square brackets in italics) and Claims (in square brackets in bold), as well as Markers \textbf{(M)}. We consider this case to be exemplary because its explanation includes reasons on why the correct answer is correct and why the incorrect answers are incorrect. We do not include argumentative relations for the sake of space and clarity.\\

\emph{QUESTION TYPE: PEDIATRICS}

\emph{CLINICAL CASE}\\

\emph{[A woman comes to the office with her 3 year old daughter because she has detected a slight mammary development since 3 months without taking any medication or any relevant history.] Indeed, [the physical examination shows a Tanner stage IV, with no growth of pubic or axillary hair.] [The external genitalia are normal.] [Ultrasonography reveals a small uterus and radiology reveals a bone age of 3 years.] What attitude should be adopted?}\\

\emph{1- [\textbf{Follow-up every 3-4 months, as this is a temporary condition that often resolves on its own.}]}

\emph{2- [\textbf{Breast biopsy.}]}

\emph{3- [\textbf{Mammography.}]}

\emph{4- [\textbf{Administration of GnRh analogues.}]}\\

\emph{CORRECT ANSWER: 1}\\

\emph{[\textbf{It seems that they want to present us with precocious puberty (or premature telarche)}] \textbf{(M)}but [they do not provide any analytical data] and [the ultrasound data are ambiguous] ([\textbf{we should assume that by a small uterus they are referring to a prepubertal uterus}], \textbf{(M)}but [they do not provide any data on ovarian size]). [We are presented with the case of a three-year-old girl with advanced mammary development, in principle without any associated cause] ([in principle she does not take drugs that can increase the level of estrogen in the blood], [she does not seem to use body creams or eat a lot of chicken meat]). [\textbf{If we follow the diagnostic scheme for a premature telarche or suspicion of precocious puberty, we request bone age and abdominal ultrasound}] ([\textbf{the EO is not advanced as in precocious puberty}, and \textbf{we assume that with a small uterus they mean a prepubertal uterus}]); [\textbf{according to the complementary examinations that we are given, it does not seem to be precocious puberty, except for the clinical (Tanner IV)}]. [\textbf{Strictly speaking, without analytical hormonal data, it seems that we could mark option 1, being necessary to follow the girl closely.}] [\textbf{If we take all the above data for granted, we could (M)rule out option 4, which would be the treatment of a central precocious puberty.}] [\textbf{Regarding the option of mammography, breast ultrasound is used in pediatrics, and in this case it would be indicated if we were told that there is breast asymmetry}] ([\textbf{we discard option 3}]). [\textbf{Regarding breast biopsy, it would only be indicated if there are warning signs.}]}

\subsection{Annotation Process and Results} \label{subsec4.2}

The annotation process consisted of three stages: training, reconciliation, and complete dataset annotation. During training, annotators worked on 10 CasiMedicos cases. We then calculated the inter-annotator agreement (IAA) results of the training phase to highlight weak spots, guideline flaws, and any issues in the dataset needing further analysis.

At the reconciliation phase, the descriptions of Claim and Premise labels were discussed and agreed upon. After this, we started the complete dataset annotation. As mentioned earlier, the original CasiMedicos dataset included 622 medical cases, but 64 cases were excluded during the annotation phase. Some of them did not have gold explanations while others were cases with confusing relations: the correct answer is a wrong disease, treatment, or diagnosis as asked in a question, thus, it is attacked by its premises instead of being supported. Therefore, the final number of annotated cases is 558. In the following subsections, we present the IAA of the entire dataset (\ref{subsec4.3}), annotation results and their description (\ref{subsec4.4}). 

\subsection{Inter-Annotator Agreement (IAA)} \label{subsec4.3}

The IAA is calculated over a random batch of 100 CasiMedicos cases. Since one instance (e.g. a claim) is usually an entire self-contained sentence, we measured the IAA at both the instance level and at the token level. In other words, we compute agreement over entire instances and over the tokens of each instance.

Table \ref{tab:IAA_instance} illustrates the IAA at instance level. Since instances are very long, annotators may be uncertain about which elements to include, leading to lower agreement scores for some labels. However, the major labels Claim and Premise have relatively good results with scores of 0.765 and 0.659, respectively. The mean F1 over all labels is 0.669.

Table \ref{tab:IAA_token} shows the IAA at token level. Here we compute the agreement over tokens of each instance. The highest agreement score is of a Claim label being 0.915, while the lowest is of a Diagnostics label accounting for 0.638. The mean F1 over all tokens is 0.880.

\begin{table}
  \centering
  \begin{tabular}{lc}
    \hline
    \textbf{Label} & \textbf{Mean F1} \\
    \hline
    Claim          &  0.765           \\
    Premise        &  0.659           \\
    Marker         &  0.642           \\
    Disease        &  0.639           \\
    Treatment      &  0.586           \\
    Diagnostics    &  0.527           \\
    \hline
  \end{tabular}
  \caption{Instance-based F1 agreement.}
  \label{tab:IAA_instance}
\end{table}

\begin{table}
  \centering
  \begin{tabular}{lc}
    \hline
    \textbf{Label} & \textbf{Mean F1} \\
    \hline
    Claim          &  0.915           \\
    Premise        &  0.891           \\
    Marker         &  0.634           \\
    Disease        &  0.738           \\
    Treatment      &  0.777           \\
    Diagnostics    &  0.638           \\
    \hline
  \end{tabular}
  \caption{Token-based F1 agreement.}
  \label{tab:IAA_token}
\end{table}

\subsection{Annotation Results} \label{subsec4.4}

In this part, we report the stats about label distribution over entire cases (documents) and the label distribution over the doctor's explanations only. Additionally, we also discuss the distribution of argumentative relations. 

Table \ref{tab:lab_distribution} reports the total number of entities over the dataset and the average number of entities per case. Table \ref{tab:distribution_explanation} shows the label distributions only for the explanations, namely, the total number of entities in explanations and the average number of entities per explanation. In both tables, we notice that the discrepancy between the average number of claims per explanation and of premises per explanation is rather high. This may seem strange since premises are needed to accept or reject claims in order to complete one argumentation unit.

However, there are plausible reasons for such distribution. First, there is a certain number of cases where the explanation is based on the evidence from the doctor's knowledge rather than clinical facts described in the case itself. Such explanations take into account the information given about the patient (e. g. age, symptoms, vital signs), but do not repeat any of these facts (as in \emph{Example 1} in Appendix \ref{Example_1}). Second, explanations that do not repeat evidence from the case are frequent, e.g. \emph{"Here we must suspect ... disease. All the symptoms fall perfectly within the picture"}; \emph{"This is a fairly easy epidemiology question, in adults without other data, Pneumococcus is the 1st"}). Last but not least, there is a group of cases with implicit premises or implicit warrants: the explanation presents claims (e.g. a conclusion about a disease and a treatment) implying that some evidences from the case text and implying certain medical knowledge to align evidences with a disease and a choice of treatment (as in \emph{Example 2} in Appendix \ref{Example_2}). 

\begin{table}
  \centering
  \begin{tabular}{lll}
    \hline
    \textbf{Label} & \textbf{Total} & \textbf{Mean per case}  \\
    \hline
    Claim          &  5021          &  8.998                  \\
    Premise        &  2313          &  4.145                  \\
    Marker         &  1117          &  2.0                    \\
    Disease        &  1791          &  3.21                   \\
    Treatment      &  1278          &  2.29                   \\
    Diagnostics    &  786           &  1.40                   \\
    \hline
  \end{tabular}
  \caption{Label Distribution over Entire Cases.}
  \label{tab:lab_distribution}
\end{table}

\begin{table}
  \centering
  \begin{tabular}{lll}
    \hline
    \textbf{Label} & \textbf{Total} & \textbf{Mean per explanation} \\
    \hline
    Claim          &  3003          &  5.948           \\
    Premise        &  470           &  0.935           \\
    Marker         &  974           &  1.833           \\
    \hline
  \end{tabular}
  \caption{Label Distribution in Explanations.}
  \label{tab:distribution_explanation}
\end{table}

In Table \ref{tab: relations} we present the distribution of argumentative relations. Support relations appear twice as much as Attack ones, making this argumentation pattern frequent and probably more convincing. In cases where the conclusion is made solely by excluding wrong propositions by attacking them, there is a lack of confidence about the claim. 

\begin{table}
  \centering
  \begin{tabular}{lll}
    \hline
    \textbf{Relation} & \textbf{Total} & \textbf{Mean per case}  \\
    \hline
    Support           &  2431          &  4.357                  \\
    Attack            &  1106          &  1.982                  \\
    \hline
  \end{tabular}
  \caption{Distribution of Argumentative Relations.}
  \label{tab: relations}
\end{table}

As a result, we present CasiMedicos-Arg, a multi-layer argument-based annotation of the English version of CasiMedicos consisting of \textbf{558} clinical cases with explanations. In the following sections, we describe the experiments performed on argument component detection (claims and premises) to establish strong baselines on the task and validate our annotations.

\section{Experimental Setup}\label{sec:experimental-setup}

We first describe the process of projecting the manually annotated argumentation labels from the source English data to the other three target languages, namely, French, Italian and Spanish. Since the annotators of the argument components were English speakers, we treated it as the source when projecting labels to the target languages. This process will result in the Multilingual Casimedicos-Arg which will then be leveraged to produce strong baselines on argument component detection using a variety of LMs, including encoders \cite{devlin-etal-2019-bert,he2021debertav3}, encoder-decoders \cite{medical-mt5} and decoder-only LLMs \cite{touvron2023llama,jiang2023mistral}.

\subsection{Multilingual CasiMedicos-Arg}

Taking the manually annotated English CasiMedicos-Arg as a starting point, we first needed to project the annotations to Spanish (original text), French and Italian (revised translations) following the method described in \citet{yeginbergenova2023cross} and \citet{yeginbergen-etal-2024-argument}. Second, and to ensure that the projection method correctly leveraged the annotations to the new data we additionally performed an automatic post-processing step of the newly generated data to correct any misalignments. Finally, to guarantee the quality of annotations and the validity of our evaluations, the translated and projected data is manually revised by native speakers. 

Label projection is performed using word alignments calculated by \textsc{awesome} \cite{dou2021word} and Easy Label Projection \cite{garcia2022model} to automatically map the word alignments into sequences (argument components) and project them from the source (English) to the target language (French, Italian and Spanish).

A particular feature of argument components is that the sequences could span over the entire length of the sentences. Therefore, after revising the automatically projected data, an extra post-processing step was performed by correcting the projections in the sequences where some annotations were placed incorrectly. The most common correction was fixing articles at the beginning of the argument components, which were systematically missed out during the automatic projection step. Other sequences were labeled only by half instead of the whole sequence. This post-processing step was essential to minimize human labor during manual correction. The number of corrections introduced during the post-processing step can be found in Appendix \ref{sec:appendix_post_processed}.

The final manual correction step involved checking the translation quality and projected labels by native expert annotators fixing any misprojections or errors in the translation. The result of this process is the Multilingual CasiMedicos-Arg dataset, obtained by projecting the manual annotations from English to Italian, French and Spanish.

\subsection{Sequence Labelling with LLMs}

We leverage Multilingual CasiMedicos-Arg to perform cross-lingual and multilingual argument component detection, a task that, due to the heterogeneity and length of the sequences, is usually a rather challenging task \cite{stab-gurevych-2017-parsing,eger-etal-2018-cross,yeginbergenova2023cross}. Furthermore, in addition to classic encoder-only models like mBERT \cite{devlin-etal-2019-bert} and mDeBERTa \citep{he2021debertav3}, we decided to also perform the task using encoder-decoder and decoder-only models. For the encoder-decoder category, we chose two variants of Medical mT5, a multilingual text-to-text model adapted to multilingual medical texts: med-mT5-large and med-mT5-large-multitask \cite{medical-mt5}. For the decoder-only architecture, we selected the LLaMa-2 \citep{touvron2023llama} and Mistral \citep{jiang2023mistral} models with 7B parameters. The domain-specific versions of these models produced less promising results, so we opted to report the results of the aforementioned models.

Previous work in sequence labeling with LLMs has demonstrated that discriminative approaches based on encoder-only models still outperform generative techniques based on LLMs \cite{wang2023gpt}. The motivation behind it is usually the nature of the sequence labeling task that even though LLMs possess some linguistic knowledge they suffer from a number of problems, notably, hallucinated content. In this paper, we use the LLMs for Sequence Labelling library to fine-tune the generative models with unconstrained decoding\footnote{\url{https://github.com/ikergarcia1996/Sequence-Labeling-LLMs}}.

We structure the experiments as follows. First, we perform \emph{monolingual} experiments in which we train and test for each language separately. Note that for English we use the gold standard annotations, while for French, Italian and Spanish we are fine-tuning the models on \emph{projected} data, which in cross-lingual transfer research is usually called \emph{data-transfer}. Additionally, we also report results of \emph{model-transfer} (fine-tuning the models in English and predicting in the rest of the target languages). Finally, we experiment with \emph{multilingual} data augmentation by pooling the training data of all four languages and then evaluating in each language separately.

Since each model has its own way of learning due to the architecture, namely, some models learn better over longer iterations and others perform at a good level in less time, we report the best results yielded from the models under different hyperparameters. Multilingual BERT and mDeBERTa were fine-tuned for 3 epochs, while Medical mT5 required 20 epochs; the rest of the hyperparameters are based on previous related work \cite{yeginbergenova2023cross} and \cite{medical-mt5}, respectively. Regarding LLaMa2 and Mistral, they were fine-tuned for 5 epochs leaving the rest of the hyperparameters as default.



\begin{table}[h]
\begin{center}
\footnotesize{
\begin{tabular}{l|cc} 
  \textbf{Model} & \textbf{Monolingual} & \textbf{Multilingual} \\
  \toprule
 mBERT           &    76.24(0.59)  &  \underline{77.14}(0.97)  \\
 mDeBERTa        &    77.08(0.89)  &  \underline{77.30}(0.59)  \\
 med-mT5-large   &    80.43(0.22)  &   \underline{82.37}(0.21) \\
 med-mT5-large-multitask  & 80.93(0.26) & \underline{82.03}(0.32) \\
 LLaMa2-7B       &    81.49(0.82)  &  \underline{83.07}(0.11)  \\
 Mistral-0.1-7B  &    \textbf{83.27}(0.48)  &  83.24(0.73) \\


 \bottomrule
\end{tabular}
\caption{F1-scores and their standard deviations for argument component detection in English CasiMedicos-Arg; \textbf{bold}: best overall result; \underline{underlined}: best result per model across the two language settings.} 
\label{table:argument_component_results_en}
}
\end{center}
\end{table}

\section{Empirical Results}\label{sec:results}


\begin{table*}
\begin{center}
\footnotesize{
\begin{tabular}{l|cccc} 
  \textbf{Model} & \textbf{Spanish} & \textbf{French} & \textbf{Italian} & \textbf{Avg.} \\
  \hline 
     & \multicolumn{3}{c}{\textbf{monolingual data-transfer}} \\
\hline
 mBERT            &   75.39(0.49)     &   73.66(0.66) &   74.78(0.59)  &   74.61 \\
 mDeBERTa         &   77.39(0.83)     &   76.35(0.29) &   76.98(0.76)  &   \underline{76.91} \\
 med-mT5-large    &   80.79(0.19)     &   80.12(0.59) &   80.32(0.04)  &   80.41 \\
 med-mT5-large-multitask & 80.69(0.65)   &   80.13(0.56) &   80.70(0.08) &   80.51 \\
 LLaMa2-7B        &   80.39(0.52)     &   80.89(0.54) &   80.69(0.46)  &   80.66  \\
 Mistral0.1-7B    &   81.71(0.29)     &   81.38(0.52) &   81.56(0.44)  &   81.55 \\

\hline
                  & \multicolumn{3}{c}{\textbf{multilingual data-transfer}} \\
\hline
 mBERT             &   75.08(0.89)     & 74.92(0.62)   &  74.95(1.38)  &  \underline{74.98}  \\
 mDeBERTa          &   76.06(1.42)     & 76.22(0.89)   &  77.06(0.65)  &  76.45 \\
 med-mT5-large     &   82.07(0.12)     & 80.85(0.26)   &  80.89(0.72)  &  \underline{81.27}  \\
 med-mT5-large-multitask & 82.09(0.26)   & 80.83(0.28)   &  80.57(0.49)  &  \underline{81.16} \\
 LLaMa2-7B         & 81.56(0.28)       & 81.03(0.49)   &  81.16(0.20)  &  \underline{81.25}  \\
 Mistral-0.1-7B    &   82.40(0.12)     & 82.10(0.33)   &  81.41(0.69)  & \underline{\textbf{81.97}}  \\
\hline
                  & \multicolumn{3}{c}{\textbf{cross-lingual model-transfer}} \\
 \hline
 mBERT                &  72.75(0.24)      & 71.47(1.27)    & 72.49(0.09)  & 72.24  \\
 mDeBERTa             &  76.05(0.14)      & 74.63(0.53)    & 75.22(0.32)  & 75.30  \\
 med-mT5-large        &  79.91(1.26)      & 78.51(1.20)    & 79.41(0.87)  & 79.28 \\
 med-mT5-large-multitask  &  79.81(0.83)    & 77.96(0.13)    & 77.07(0.34)  & 78.28    \\
 LLaMa2-7B            &  75.31(0.68)      & 68.56(1.07)    & 73.86(0.51)  & 72.58   \\
 Mistral-0.1-7B       &  79.27(0.42)      & 70.62(7.37)    & 78.36(0.37)  & 76.08  \\
 \bottomrule
\end{tabular}
\caption{F1-scores and their standard deviations of data-transfer (monolingual and multilingual), and cross-lingual model-transfer experiments using Spanish, French, and Italian data; \textbf{bold}: best overall result; \underline{underlined}: best result per model across the three language settings.} 
\label{table:argument_component_results}
}
\end{center}
\end{table*}

In this section we report the results obtained after performing the steps described in Section \ref{sec:experimental-setup}. All the results and standard deviations reported in this section are obtained by averaging three randomly initialized runs. We evaluate using sequence level F1-macro score, a common metric for argument component detection.

We first show the results on monolingual (using the manually annotated English data) and multilingual (fine-tuning on all four languages and evaluating in English) in Table \ref{table:argument_component_results_en}. Overall, it can be observed that the decoder-only generative models outperform the rest, though the Medical mT5 models are nearly as effective. Furthermore, the \emph{multilingual} method of pooling all languages into a single dataset proves to be beneficial for every model, improving over the results obtained when training using the gold standard English data only.

The results for Spanish, French and Italian are displayed in Table \ref{table:argument_component_results}. As for the English results, it can be seen that the \emph{multilingual data-transfer} approach is the most effective setting, even with LLMs which are supposedly pre-trained on English data only. Among all the models, Mistral achieves the highest F1-macro scores. However, while for all the other models the multilingual training was advantageous no substantial improvement was observed in a similar setting with Mistral. Finally, it can be seen that \emph{cross-lingual model transfer} is the least optimal of the settings, even when using state-of-the-art multilingual LMs such as mDeBERTa \cite{he2021debertav3}. An interesting point to note is that for \emph{cross-lingual model transfer} the best results are obtained by the Medical mT5 models, which may be due to this model being trained on multilingual medical data \cite{medical-mt5}.

Summarizing, in this section we present competitive baselines for argument component detection on CasiMedicos-Arg, validating both the manual annotations and the strategy of projecting English labels to other languages to facilitate the application of cross-lingual and multilingual techniques.

\section{Conclusion} \label{sec6}

In this paper, we present CasiMedicos-Arg, a multilingual (French, English, Italian and Spanish) Medical QA dataset including gold reference explanations written by medical doctors which has been annotated with argumentative structures. This dataset aims to bridge a glaring gap in the Medical QA ecosystem by facilitating the evaluation of explanations generated to argue or justify a given prediction.

The final dataset includes 558 documents (parallel in four languages) with reference gold doctors' explanations which are enriched with manual annotations for argument components (5021 claims and 2313 premises) and relations (2431 support and 1106 attack).

Both inter-annotator agreement results and the baselines provided for argument component detection demonstrate the validity of our annotations. Furthermore, experiments show the advantage of performing argument component detection from a \emph{multilingual data-transfer} perspective.

\section*{Limitations}

We consider two main limitations in our work that we would like to address in the short term future. First, the choice of languages. We would have liked to include languages from different language families and with different morphological and grammatical characteristics, but we were limited by the native expertise available to us to perform the manual corrections of the projected labels and translations. Second, the size of the dataset (558 documents) could be larger. 

Regarding the first limitation, we still think that our experiments demonstrate the superiority of performing \emph{multilingual data-transfer} over \emph{cross-lingual model transfer}, at least with the LLMs currently available. With respect to the size of the dataset, we would like to point out that its size is similar to other datasets reviewed in Section \ref{sec2}, which are being widely used to benchmark LLMs for Medical QA.

Another issue worth considering in the future is the need to further research the generation of explanations for the predictions while taking into account a crucial unsolved issue, namely, the evaluation explanation generation in the highly specialized medical domain.

\section*{Acknowledgments}
We thank the CasiMedicos Proyecto MIR 2.0 for their permission to share their
data for research purposes. This work has been supported by the French government, through the 3IA Côte d'Azur Investments in the Future project managed by the National Research Agency (ANR) with the reference number ANR-19-P3IA-0002. This work has also been supported by the CHIST-ERA grant of the Call XAI 2019 of the ANR with the grant number Project-ANR-21-CHR4-0002.
We are also thankful to several MCIN/AEI/10.13039/501100011033 projects: (i) Antidote (PCI2020-120717-2), and by European Union NextGenerationEU/PRTR; (ii) DeepKnowledge (PID2021-127777OB-C21) and ERDF A way of making Europe; (iii)  DeepMinor (CNS2023-144375) and European Union NextGenerationEU/PRTR. We also thank the European High Performance Computing Joint Undertaking (EuroHPC Joint Undertaking, EXT-2023E01-013) for the GPU hours. Anar Yeginbergen's PhD contract is part of the PRE2022-105620 grant, financed by MCIN/AEI/10.13039/501100011033 and by the FSE+.

\bibliography{custom}

\clearpage

\appendix

\section{Appendix. CasiMedicos Real Cases} \label{sec:appendix_real_cases}

\emph{\textbf{Example 1:}} \label{Example_1}\\

\emph{QUESTION TYPE: DERMATOLOGY}

\emph{CLINICAL CASE:}\\

\emph{A 62-year-old man with a history of significant alcohol abuse, carrier of hepatitis C virus, treated with Ibuprofen for tendinitis of the right shoulder, goes to his dermatologist because after spending two weeks on vacation at the beach he notices the appearance of tense blisters on the dorsum of his hands. On examination, in addition to localization and slight malar hypertrichosis.  The most likely diagnosis is:}\\

\emph{1- Epidermolysis bullosa acquisita.}

\emph{2- Porphyria cutanea tarda.}

\emph{3- Phototoxic reaction.}

\emph{4- Contact dermatitis.}

\emph{5- Acute intermittent porphyria.}\\

\emph{CORRECT ANSWER: 2}\\

\emph{Porphyria Cutanea Tarda: 60\% of patients with PCT are male, many of them drink alcohol in excess, women who develop it are usually treated with drugs containing estrogens. Most are males with signs of iron overload, this overload reduces the activity of the enzyme uroporphyrinogen decarboxylase, which leads to the elevation of uroporphyrins. HCV and HIV infections have been implicated in the precipitation of acquired PCT. There is a hereditary form with AD pattern. Patients with PCT present with blistering of photoexposed skin, most frequently on the dorsum of the hands and scalp. In addition to fragility, they may develop hypertrichosis, hyperpigmentation, cicatricial alopecia and sclerodermal induration.}\\

\emph{\textbf{Example 2:}} \label{Example_2}\\

\emph{QUESTION TYPE: PEDIATRICS}

\emph{CLINICAL CASE:}\\

\emph{6-month-old infant presenting to the emergency department for respiratory distress. Examination: axillary temperature 37.2°C, respiratory rate 40 rpm, heart rate 160 bpm, blood pressure 90/45 mmHg, SatO2 95\% on room air. He shows moderate respiratory distress with intercostal and subcostal retraction. Pulmonary auscultation: scattered expiratory rhonchi, elongated expiration and slight decrease in air entry in both lung fields. Cardiac auscultation: no murmurs. It is decided to keep the patient under observation in the hospital for a few hours. What do you consider the most appropriate attitude at this time with regard to the complementary tests?}\\

\emph{1- Request venous blood gas, leukocyte count and acute phase reactants.}

\emph{2- Request chest X-ray.}

\emph{3- Request arterial blood gases and acute phase reactants.}

\emph{4- Do not request complementary tests.}\\

\emph{CORRECT ANSWER: 4}\\

\emph{The patient probably presents with bronchiolitis. At this stage, no additional tests should be performed unless there is a clinical worsening.}

\section{Number of corrections after annotation projection}\label{sec:appendix_post_processed}
The number of corrections required after automatically projecting the annotations.
\begin{table}[h]
  \centering
  \begin{tabular}{c|c}
    \hline
    \textbf{Set (Language)} & Number of corrections       \\
    \hline
        Train (ES)             &      450                 \\
        Test (ES)              &      153                 \\
        Dev (ES)               &      64                  \\

    \hline
        Train (FR)             &     378                  \\
        Test (FR)              &     109                  \\
        Dev (FR)               &     49                   \\

    \hline
        Train (IT)             &     336                  \\
        Test (IT)              &     117                  \\
        Dev (IT)               &     55                   \\
    \hline
  \end{tabular}
  \caption{Number of corrections introduced in the post-processing step after automatic label projection.}
  \label{tab: post_processed}
\end{table}

\end{document}